\documentclass[lettersize,journal]{IEEEtran}

\IEEEoverridecommandlockouts   

\usepackage{cite}
\usepackage{amsmath,amssymb,amsfonts}
\usepackage{algorithmic}
\usepackage{graphicx}
\usepackage{textcomp}
\usepackage{xcolor}
\usepackage{tikz}
\usepackage{pgfplots}
\usepackage{subcaption}
\usepackage{booktabs}
\usepackage{adjustbox}
\usepackage{rotating}
\usepackage{float}
\usepackage{geometry}
\usepackage[caption=false,font=normalsize,labelfont=sf,textfont=sf]{subfig}
\usepackage{pgfplots}
\pgfplotsset{compat=1.18}

\begin{document}

\title{Torque Responsive Metamaterials Enable High Payload Soft Robot Arms}

\author{Ian Good, Srivatsan Balaji, David Oh, Sawyer Thomas, and Jeffrey I. Lipton% <-this % stops a space
%\thanks{*Co-First Authors}%
\thanks{This work was supported by the National Science Foundation, grant numbers 2017927 and 2035717, by the ONR through Grant DB2240 and by the Murdock Charitable Trust through grant 201913596. (Corresponding Author: Jeffrey I. Lipton.)}% <-this % stops a space
\thanks{Ian Good, Srivatsan Balaji, David Oh, and Sawyer Thomas are with the Mechanical Engineering Department, University of Washington, Seattle, WA 98195 USA} %(e-mail: iangood@uw.edu; sri.vatsan0307@gmail.com).}%
\thanks{Jeffrey Ian Lipton is with
the College of Engineering, Mechanical and Industrial Engineering, Northeastern University, Boston, MA 02115 USA (e-mail: j.lipton@northeastern.edu).}}%

\maketitle
\thispagestyle{empty}
\pagestyle{empty}

\begin{abstract}
Soft robots have struggled to support large forces and moments while also supporting their own weight against gravity. This limits their ability to reach certain configurations necessary for tasks such as inspection and pushing object up.
We have overcome this limitation by creating an electrically driven metamaterial soft arm using handed shearing auxetics (HSA) and bendable extendable torque resistant (BETR) shafts. These use the large force and torque capacity of HSAs and the nestable torque transmission of BETRs to create a strong soft arm.
We found that the HSA arm was able to push 2.3 kg vertically and lift more than 600 g when positioned horizontally, supporting 0.33 Nm of torque at the base. The arm is able to move between waypoints while carrying the large payload and demonstrates consistent movement with path variance below 5 mm. The HSA arm's ability to perform active grasping with HSA grippers was also demonstrated, requiring 20 N of pull force to dislodge the object. Finally, we test the arm in a pipe inspection task. The arm is able to locate all the defects while sliding against the inner surface of the pipe, demonstrating its compliance.
\end{abstract}

%%%%%%%%%%%%%%%%%%%%%%%%%%%%%%%%%%%%%%%%%%%%%%%%%%%%%%%%%%%%%%%%%%%%%%%%%%%%%%%%%%%%%%%%%%%%%%%%%%%%%%%%%%%%%%%%%%%%
%%%%%%%%%%%%%%%%%%%%%%%%%%%%%%%%%%%%%%%%%%%%%%%%%%%%%%%%%%%%%%%%%%%%%%%%%%%%%%%%%%%%%%%%%%%%%%%%%%%%%%%%%%%%%%%%%%%%
%Link to Whitesides Draft:
%https://docs.google.com/document/d/1W4OIRnaOT9GxNQZPAdxWpscXf8zlMW68PIzMT95d5G8/edit?usp=sharing

\section{Introduction}
\begin{figure}[t]
    \centering
    %\hfill
    \includegraphics[width=1\linewidth]{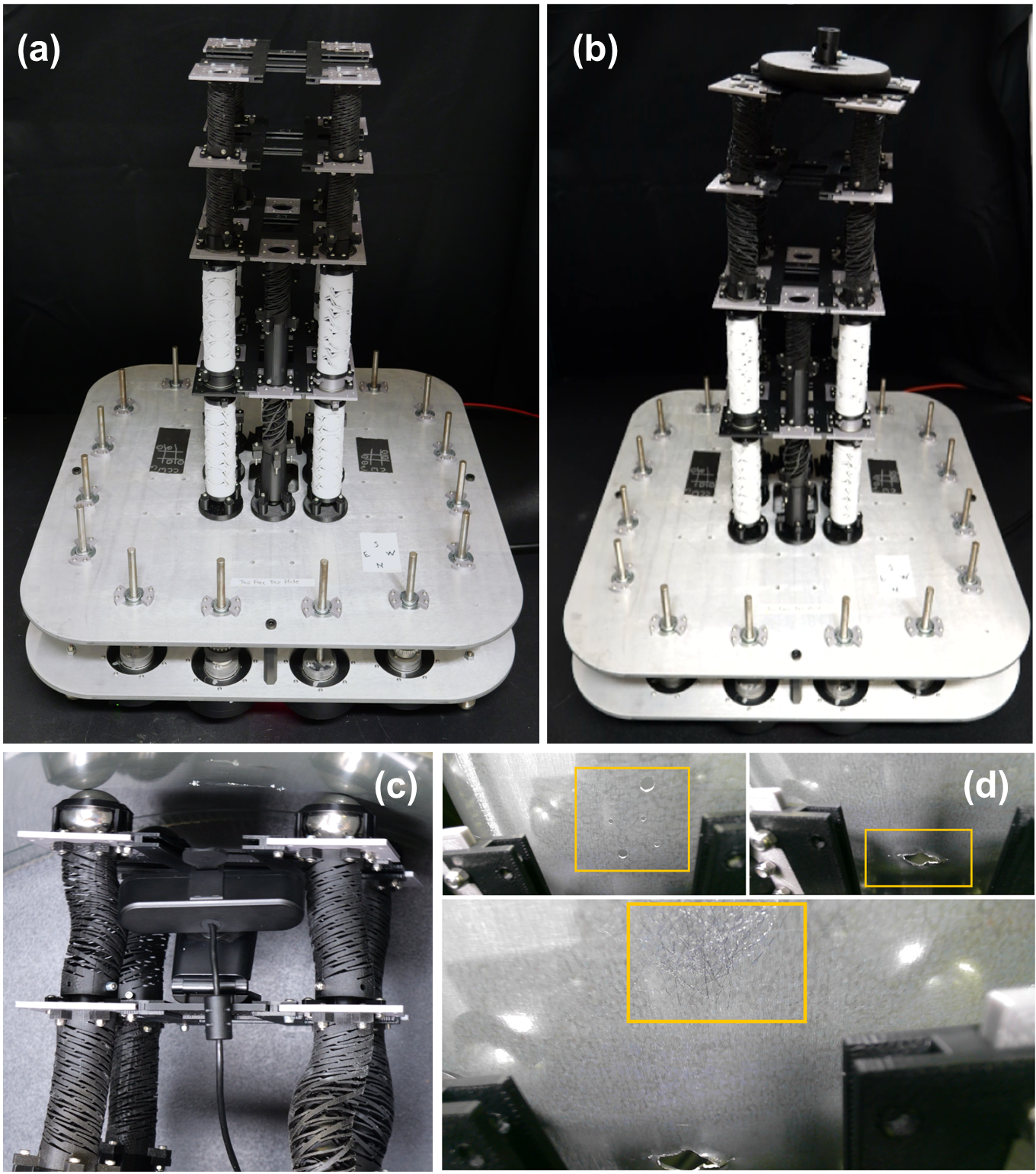}
    %\captionsetup{justification=centering}
    \caption{We present a soft Handed Shearing Auxetics (HSA) robot arm that is able to support meaningful forces and torques while holding itself up against gravity. Subfigure (a) shows the assembled soft HSA arm, (b) shows the soft HSA arm lifting a 2.3 kg payload vertically, (c) shows the inspection task setup with the mounted webcam, and (d) shows the different pipe defects (holes, gash, scratches from top left) captured by the HSA arm during the inspection task.}
    \label{fig:heroArm}
    %\vspace{-0.5cm}
\end{figure}

Soft robot arms have been trapped in hanging form factors and low payloads due to their inability to support larger forces and moments. Hanging form factors restrict a soft robot's ability to interact with their environment to a top-down perspective, rather than the one that most benefits the task at hand. We present an electrically driven metamaterial robot arm that is capable of lifting large payloads while also supporting its full weight under gravity as seen in Fig. \ref{fig:heroArm} (a). This allows the arm to operate in whatever orientation best serves the current task.

Soft robot arms cover a broad area of research\cite{yasa2023overview,stella2023SoftDesignReview}. They have been used to explore ruins\cite{coad2019vine}, for sample collection underwater\cite{xie2023octopus}, for screw assembly tasks \cite{carton2024HSASphereArm}, and even mounted on drones for docking\cite{ahlquist2024DroneArm}. A major benefit of soft robot arms comes from their distributed compliance which helps provide a physical guarantee of safety and with tasks such as pick and place or peg in hole. However, they are broadly held back by two limitations: small resisted torques and payload capacities.

Supporting large forces and torques is critical for robot arms. Li et al \cite{li2017fluid} were able to create tremendous force capacities through fluid-driven artificial muscles, capable of lifting car tires. However these are limited to pulling through the body of the soft arm and can not support meaningful torques. The combination of both meaningful force and torque resistance is vital to a useful robot arm. We build a soft robotic arm that is capable of lifting large payloads while supporting its own weight under gravity.

%hero figure
%% Layer Overview
\begin{figure*} [h]
    \centering
    %\hfill
    \includegraphics[width=1\linewidth]{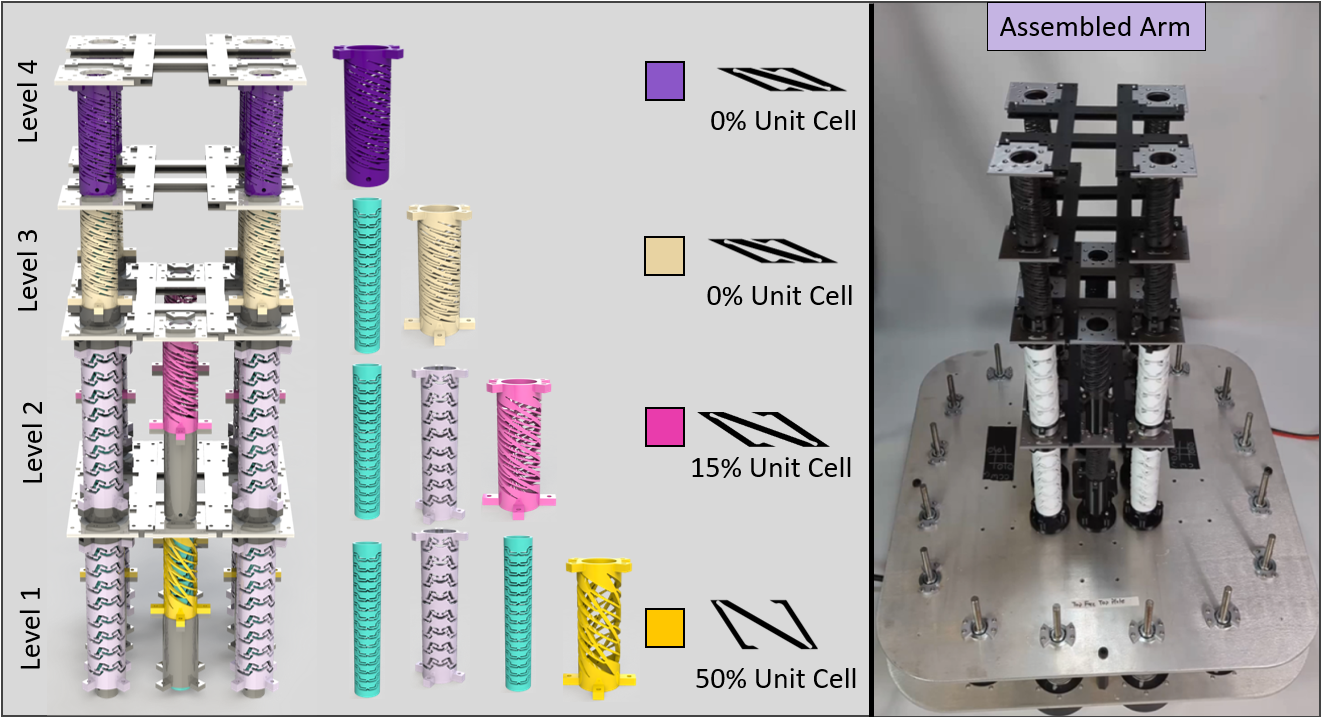}
    %\captionsetup{justification=centering}
    \caption[Soft Robot Component Overview]{A System overview is presented for the HSA Arm. The arm is made from four layers, with each driven by a different HSA. Proximal layers use HSAs printed further along the auxetic trajectory culminating in a thinner distal HSA. HSAs on level two through four are driven by nested Bendable Extendable Torsionally Rigid (BETR) shafts. A rendered version of the arm is shown on the left and the realized construction can be seen on the right.}
    \label{fig:SystemLayerOverview}
    \vspace{-0.2cm}
\end{figure*}

In this paper we:
\begin{itemize}
    \item Design a soft robot arm using handed shearing auxetics capable of lifting more than 2 kg vertically and 600 g horizontally.
    \item Analyze the design space of handed shearing auxetics and characterize bendable, extendable flex shafts
    \item Evaluate the arm through object lifting, docking and pipe inspection tasks
\end{itemize}

\section{Background}

There is a large body of recent literature on soft robot arms. Guan et al. created an arm capable of lifting strawberries through cable-driven helicoid structures\cite{guan2023trimmed}. Kang et al. and Zhao et al. used shape memory alloy muscles to lift water cups and ping pong balls\cite{kang2023shapeWaterCup,zhao2023starblocksModular}. Ahlquist et al., Lui et al. and Bianchi et al. all created pneumatic arms to dock, pick flowers, and throw ping pong balls \cite{ahlquist2024DroneArm,liu2022FlowerArm,bianchi2023softoss}. All of these works have two things in common: Small payloads, and an inability to support themselves outside of a hanging configuration. These arms cannot be inverted and would buckle while trying to support their own weight plus their payload. These are major restrictions. The field of robotics has centralized on objects from the YCB dataset \cite{YCBObjects} which have masses much greater than that of a ping pong ball or strawberry. To better leverage the benefit of soft robots, we need to support payloads in the hundreds of grams range. 

Some soft robot arms can lift significant payloads. Li et al. used pnematic muscles to lift a car tire \cite{li2017fluid}. While capable of supporting meaningful payloads, this arm cannot support itself outside of the trunk configuration and can only support large forces, not moments. Childs et al.,Yang et al., and Bruder et al. have created soft robotic arms capable of meaningful payloads while supporting moments; however, they support forces and torques smaller than the work presented here \cite{childs2021rhombus,yang2024Heavy-Load,bruder2023increasing}.

Handed Shearing Auxetic (HSA) soft robots have emerged as a promising field for their ability to generate large forces and for their ability to hold themselves up against gravity \cite{lipton2018handedness, good2021expanding}. Work has been done to generate controls and models for 4-DoF platforms resisting gravity \cite{garg2022kinematic,stolzle2023modelling}. They have also been used in mobile robots \cite{ketchum2023TurtleBot,kim2024Inchworm}, in packing groceries \cite{chen2024GroceryPacking}, and in manipulating YCB objects \cite{good2024trsll}. For mobile platforms, payloads are still relatively small, and for the TRSLL and grocery packing robots, only the grippers were soft. The largest limitation however is that handed shearing auxetic robots are limited to a single layer of HSA, preventing their use as a robot arm. Recent works \cite{Thomas2025BETR,kim2024Inchworm} have enabled torque transmission to HSAs not directly connected to a motor.

This points towards electrically driven soft robots as a capable solution for soft robot arms. HSAs can support large forces and torques while supporting themselves under gravity and now no longer require a direct connection to a motor to actuate.

\section{Design of Metamaterial Soft Robot Arm}

\begin{figure*}
    \centering
    \includegraphics[width=0.99\textwidth]{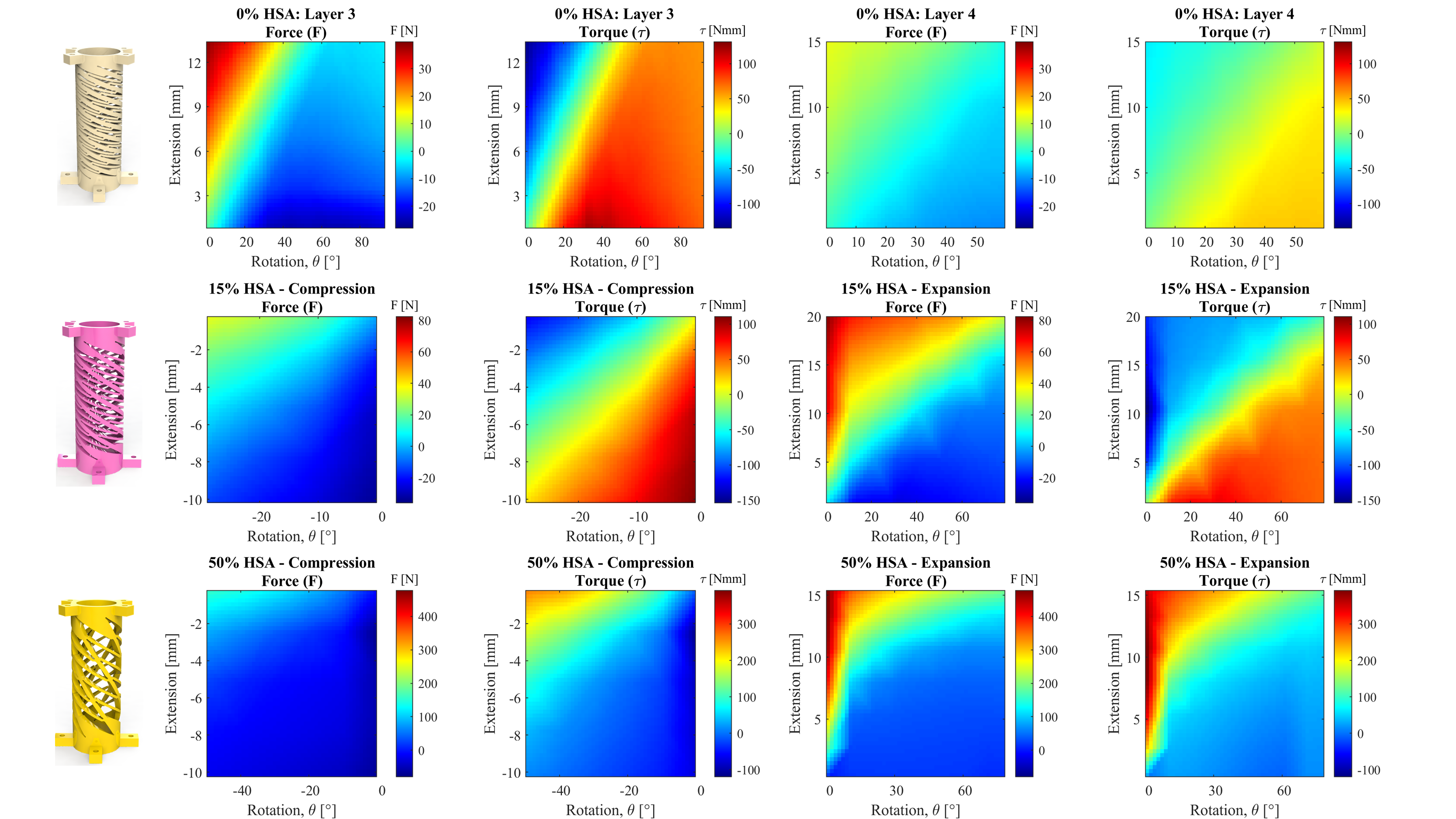}
    \caption[HSA Characterization]{Interpolated HSA force and torque characterization is presented here for each of the four layers of HSA in the soft arm. Thick-  and thin-walled HSAs printed at 0\% of way along the auxetic trajectory are presented with just expansion data (layer 3, 4 respectively) and HSAs printed beyond that (layer 1, 2) are shown with both expansion and compression regions.}
    \label{fig:HSAARMCharacterization}
    %\vspace{-0.4cm}
\end{figure*}
\begin{table*}[]
\centering
\caption{Table showing the minimum and maximum force (F) and torque ($\tau$) values for all the HSA configurations from the mechanical characterization data. Here, $\theta$ represents the corresponding value of rotation and \textbf{e} represents the corresponding value of extension. The table is divided into two categories to represent the values in the compression and expansion regimes.}
\label{tab:HSACharacterizationTable}
\setlength{\tabcolsep}{20pt}
\renewcommand{\arraystretch}{1.1}
\begin{tabular}{ccccc}
\multicolumn{5}{c}{\textbf{COMPRESSION}}                                                                                  \\ \hline
\textbf{HSA Type} &
  \textbf{\begin{tabular}[c]{@{}c@{}}Min. Force ($\theta$, e)\\ {[}N{]}\end{tabular}} &
  \textbf{\begin{tabular}[c]{@{}c@{}}Max. Force ($\theta$, e)\\ {[}N{]}\end{tabular}} &
  \textbf{\begin{tabular}[c]{@{}c@{}}Min. Torque ($\theta$, e)\\ {[}Nmm{]}\end{tabular}} &
  \textbf{\begin{tabular}[c]{@{}c@{}}Max. Torque ($\theta$, e)\\ {[}Nmm{]}\end{tabular}} \\ \hline
50\%                 & -79.5 (0, -2.6)      & 167.3 (-49.9, 0)     & -120.3 (0, -2.6)     & 275.8 (-49.9, 0)     \\
15\%                 & -34.5 (0, -10.4)     & 38.3 (-29.9, 0)      & -130.2 (-29.9, 0)    & 110.89 (0, -10.4)    \\
0\% (Thick)          & -                    & -                    & -                    & -                    \\
0\% (Thin)           & -                    & -                    & -                    & -                    \\ \hline
\\
\multicolumn{5}{c}{\textbf{EXPANSION}}                                                                                    \\ \hline
\textbf{HSA Type} &
  \textbf{\begin{tabular}[c]{@{}c@{}}Min. Force ($\theta$, e)\\ {[}N{]}\end{tabular}} &
  \textbf{\begin{tabular}[c]{@{}c@{}}Max. Force ($\theta$, e)\\ {[}N{]}\end{tabular}} &
  \textbf{\begin{tabular}[c]{@{}c@{}}Min. Torque ($\theta$, e)\\ {[}Nmm{]}\end{tabular}} &
  \textbf{\begin{tabular}[c]{@{}c@{}}Max. Torque ($\theta$, e)\\ {[}Nmm{]}\end{tabular}} \\ \hline
50\%                 & 0 (0, 0)             & 478.1 (0, 15.7)      & -8.8 (59.9, 0)       & 392.2 (0, 10.4)      \\
15\%                 & -26.3 (29.9, 0)      & 82.5 (0, 20.4)       & -153.29 (0, 10.2)    & 87.9 (9.9, 20.4)     \\
0\% (Thick)          & -27.9 (42, 0)        & 39.7 (0, 13.6)       & -134.8 (0, 13.6)     & 131 (31.3, 0)        \\
0\% (Thin)           & -11.21 (59.9, 0)     & 13.1 (0, 15.3)       & -40.2 (0, 15.3)      & 47 (19.9, 0)         \\ \hline
\end{tabular}
\end{table*}
This section describes the design of our four layer soft metamaterial robot arm. First, we use a lumped element model to determine component performance targets for horizontal payloads of 500 g and 2 kg vertically. Then, we manufacture and test components for both extension performance and torque transmission. We finally evaluate the assembled system to see how well it meets the initial design requirements.

\subsection{HSA Characterization}
%% Component Characterization

Using kinematics, we determined force and torque targets corresponding to a 2 kg vertical lift and a 500 g horizontal lift. Based on work from \cite{good2021expanding,truby2021recipe}, we designed HSAs to achieve the target properties, then manufactured them as described in \ref{characterization}.

We characterize HSAs using two main properties, force (F) and torque ($\tau$) as a function of both displacement and rotation. This extends the characterization from \cite{good2021expanding,truby2021recipe} by characterizing properties beyond the minimum energy state. We characterize four different configurations of the HSA, 50\% open, 15\% open, 0\% open, and 0\% open with thinner walls. These represent the four layers of the arm. The 50\% open HSAs occupied the first layer of the arm, 15\% open HSAs occupied the second layer, and the 0\% HSAs occupied layers three and four as shown in Fig. \ref{fig:SystemLayerOverview}. For layer 4, we reduced the wall thickness of the HSA to reduce the torque requirement for actuation. All of the HSAs on layers 1, 2 and 3 (50\%, 15\%, 0\%) had a length of 90.5 mm, an outer diameter of 30.7 mm, and a wall thickness of 3.5 mm. Layer 4 HSAs (0\%) had the same length and outer diameter with a reduced wall thickness of 1.7 mm.

\begin{figure*}
    \centering
    \includegraphics[width=1\textwidth]{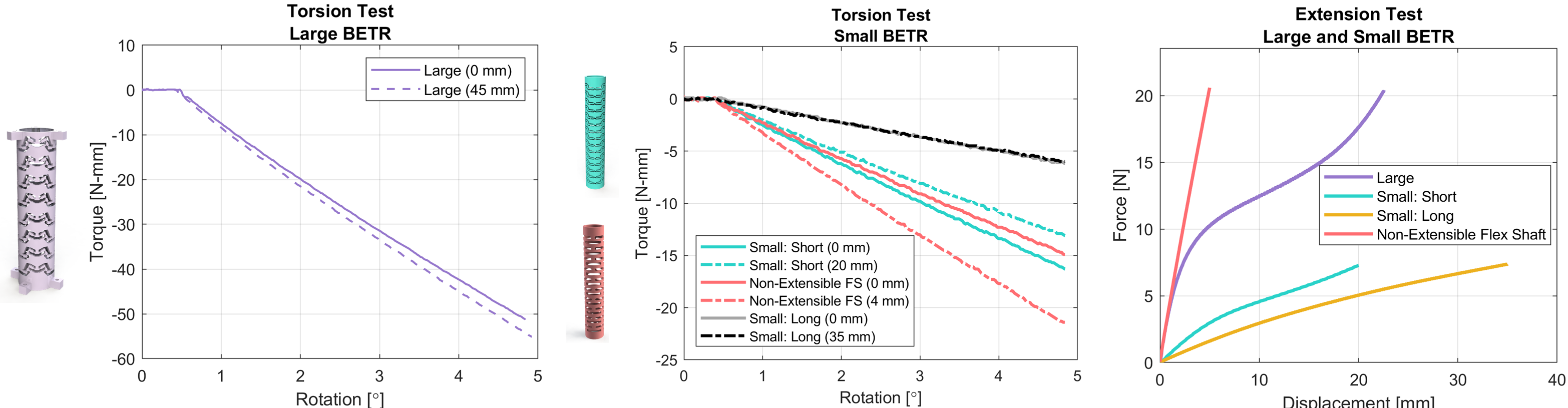}
    \caption[Torque Transmission Characterization]{Plot of torsion and extension tests for the soft soft torque transmission elements in the robot arm. We characterize elements for layer 2 (Small: Short), layer 3 (Large BETR), and layer 4 (Small: Long). We also characterize the single non-extendable flex shaft for complete individual component level analysis. Torque characterization was done when in the extended (dashed) and rest states (solid). The X intercept represents the slack/deadband in the material and the slope represents the torsional stiffness. Forces required to extend the elements in the arm are also shown in the extension test plot. These values oppose HSA extension.
    }
    \label{fig:BETRCharacterization}
    \vspace{-0.2cm}
\end{figure*}

\subsection{Transmission Characterization} \label{characterization}

All HSAs we characterized were fabricated using the Carbon M1 3D printer with FPU 50. All HSAs were printed horizontally on the build platform, and were post processed following the manufacturer's guidelines and specifications.

The test procedure consisted of a series of extensions and rotations that were programmed using a custom test method on an Instron 68SC-2 at 200 Hz. We individually chose the extension and rotation values based on the maximum safe limits of the tested HSA configurations. The ranges of extension and rotation were determined by manually observing them for all HSA configurations. The data was collected over seven and fourteen equidistantly spaced points in displacement and rotation respectively. This results in 98 datapoint groups per HSA. Each datapoint group was collected during a 1 mm vertical displacement at 10 mm/s. The raw data collected from the test apparatus was approximated using cubic fits for displacement and rotation to visualize the force (F) and torque ($\tau$) responses of the HSA as shown in Fig. \ref{fig:HSAARMCharacterization}.

A linear interpolation surface fit was chosen to visualize the data. For the force plot, positive values represent the HSA pulling on the environment while negative values represent pushing on the environment. For torque, positive values represent the HSA trying to twist less in the direction it is twisting and negative values represent the HSA trying to twist more than the direction it is twisting in. The maximum and minimum force and torque values observed for each HSA configuration is shown in Table \ref{tab:HSACharacterizationTable}.

To deliver torque to layers two through four, a bendable, extendable torque transmission system is needed. We looked to the work presented in \cite{Thomas2025BETR}. We characterized three different configurations of the extendable flex shafts (BETRs), large, long, and short. We characterized their force and torque responses using an Instron 68SC-2 mechanical testing machine. As these BETRs are intended for transmitting torque from the motors to the HSAs, we evaluated the torsional stiffness of these BETRs, while also testing the force required to extend them. Additionally, we incorporated a non-extensible flex shaft at the first layer of the arm to provide enhanced torsional stiffness, while the extendable BETRs were used for the second and third layer. The CAD rendering of the BETR designs can be seen in Fig. \ref{fig:SystemLayerOverview}. The large BETR had an outer diameter of 30.7 mm and a wall thickness of 4.5 mm. They were fabricated with PTFE tubes using a Bantam Desktop CNC machine. The large BETRs were utilized to drive the arm's third layer. Additionally, we fabricated two variants of small BETRs, short and long. The small BETRs were made out of FPU 50 using a Carbon M1 3D printer, and can be seen in Fig. \ref{fig:BETRCharacterization}. The long BETR was fabricated by connecting two short BETRs and a single non-extendable flex shaft using 3D-printed adaptors made out of PLA. The short BETRs were used to drive the 15\% HSAs on the second layer of the HSA arm, while the long BETRs were used to drive the 0\% HSAs on layer 4 as shown in Fig. \ref{fig:SystemLayerOverview}.

The torque response was recorded for the large, short, and long BETRs and the non-extendable flex shaft by applying a rotation of 5 degrees at a rate of 1 deg/s. This characterization was done in both non-extended and extended states for each of the components to understand how torsional stiffness changes with extension. Little change was observed for the BETRs. The large, long, and short BETRs were extended to 45 mm, 35 mm, and 20 mm respectively, while the non-extendable flex shaft was extended to 4 mm. Similarly, for force response, an extension of 22 mm, 20 mm, 35 mm and 5 mm was used for the large, short, long, and non-extensible flex shaft respectively at the rate of 1 mm/s. All of the recorded data were visualized and can be seen in Fig. \ref{fig:BETRCharacterization}.

The torsional stiffness values were then calculated from raw data for each of the BETRs. The large BETR had torsional stiffness values of -11.52 Nmm/deg and -12.03 Nmm/deg at the non-extended and extended states respectively. The long BETR was at -1.39 Nmm/deg and -1.33 Nmm/deg at the non-extended and extended states, showing very little difference in its performance. The short BETR had torsional stiffness values of -3.60 Nmm/deg and -2.90 Nmm/deg at its non-extended and extended states. The non-extendable flex shaft had torsional stiffness values of -3.29 Nmm/deg and -4.78 Nmm/deg at the non-extended and extended states respectively.

\subsection{Fabrication of the Soft Arm} \label{Fabrication of Soft Arm}
In this section, we describe the manufacturing and characterization of the Handed Shearing Auxetic (HSA) actuators, and Bendable Extendable Torsionally-Rigid (BETR) flex shafts. We also describe the fabrication and assembly steps for the HSA arm. We then describe the test methods that were used to evaluate the performance of the robot arm.

A parallel plate system was used to connect the sixteen independent Quasi-Direct Drive motors (QDD100 Beta 3) to the HSAs and BETRs through pulleys. Two torques were transmitted along the same shaft using a set of live and dead axle pulleys with the motors preventing unwanted cross-talk. This forms the metal base at the bottom of the robot that the soft arm is built on top of. 

The components used for the fabrication of the robot arm can be seen in Fig. \ref{fig:SystemLayerOverview}. The HSA arm is made from four levels or layers, where each level was connected to the other layer using 3D-printed plate sections.

For the first level, we added four large BETRs (lavender) with one in each section that drives another four large BETRs at level 2. The large BETRs drive the 0\% HSA (gold) at level 3. Long BETRs (teal) run through the inner sections of the large BETR up to level 4 to drive the thinner HSAs (purple) at the distal end. Levels 1 and 2 are driven using 50\% (yellow) and 15\% (pink) HSAs at levels 1 and 2 respectively. In order to maximize force output, the HSAs were connected to 3D-printed adaptors at levels 1 and 2 as shown in gray in Fig. \ref{fig:SystemLayerOverview}.
\section{Results - Performance Evaluation of HSA Arm} 

In this section, we evaluate the performance of our HSA arm. The arm performs lifting tasks to demonstrate payload capabilities in both a horizontal and vertical configuration. The soft HSA arm's active grasping capability was evaluated by measuring the pull force required to dislodge an object from its grasp. This test aims to demonstrates the arm's ability to perform tasks like underwater docking, where significant forces are experienced in the lateral and longitudinal directions due to ocean environments. The arm's compliance is evaluated by inspecting a pipe for defects, a common application in the Oil and Gas industry. The workspace of the arm is evaluated using an optical motion tracking system. This demonstrates the HSA arm's reachable positions in 3D space. The arm was then tested for its repeatability by programming it to follow triangular and square patterns for a fixed number of cycles.

\subsection{Payload Evaluation}
We demonstrate the payload capacity of the metamaterial soft robot arm in both horizontal (extending out) and vertical (extending up) configurations. 

\begin{figure}[t]
    \centering
    \includegraphics[width=\columnwidth]{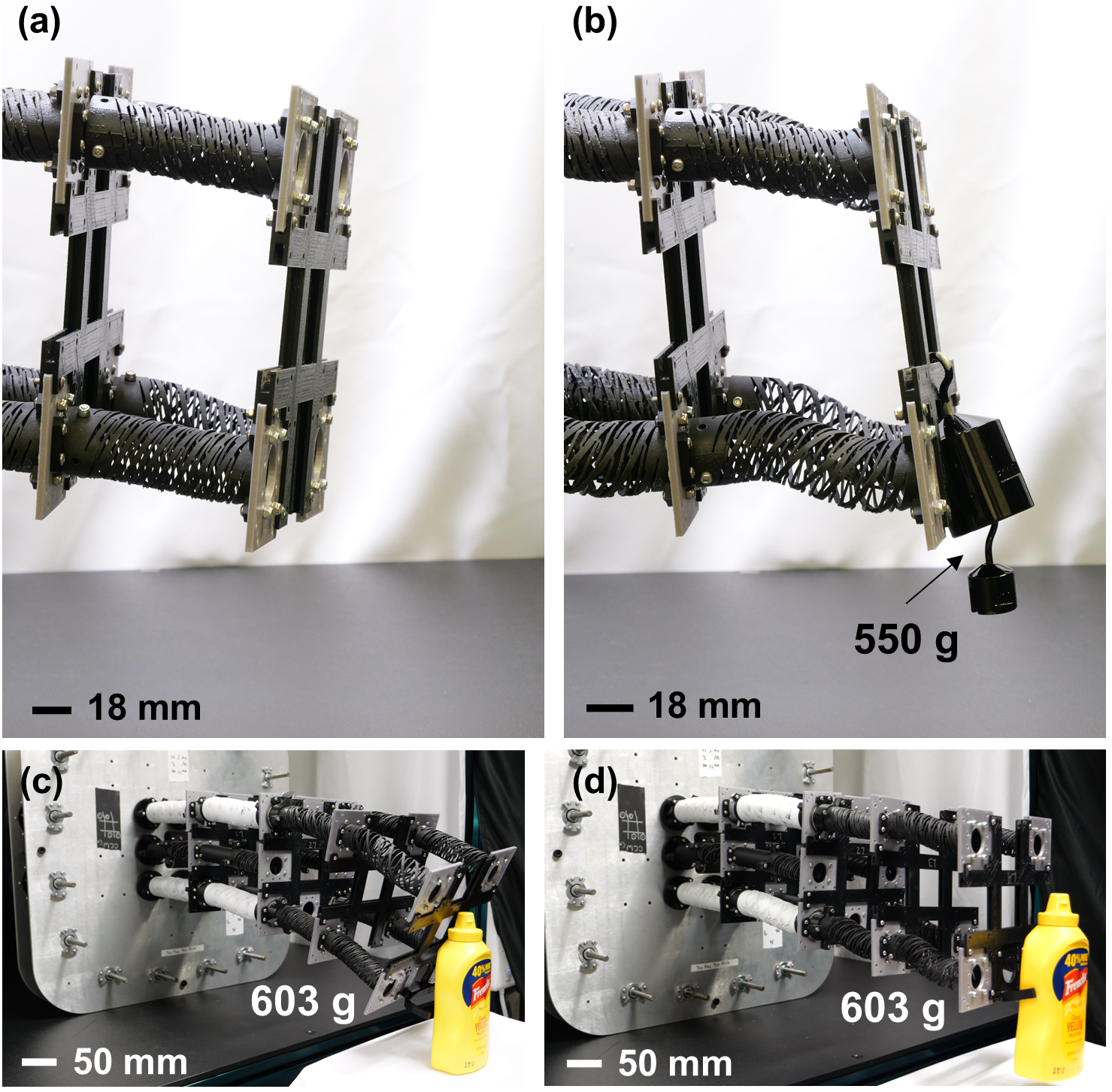}
    \caption[Horizontal Payload Test]{Here we present the maximum load supported when the arm is positioned horizontally (b). First the arm's default resting position is recorded with a laser level. Then the arm is moved up to its maximum extension. Weights are added to the distal tip until the arm returns to its starting position. The arm is able to support 600 g when horizontally extended, resulting in 0.33 Nm of resisted torque. Subfigures (c) and (d) show the HSA arm before and after lifting a 603-g mustard bottle from the YCB object dataset.}
    \label{fig:horizontalPayload}
    %\vspace{-0.3cm}
\end{figure}

The horizontal test is conducted following the methods established in \cite{yang2024Heavy-Load}. The arm is commanded to move up against gravity from the home position and weights are added until it returns home. The home position was recorded as breaking the beam of the laser and the test ended when the beam was broken again. The arm was able to support 550g in weights without breaking the beam but failed to support 575g. The weights were suspended 60 cm out from the base, resulting in a torque to the base of the arm of 0.33 Nm. This demonstrates a huge payload, especially for a soft robot arm that is supporting itself perpendicular to gravity.

A vertical lift test was also conducted as shown in Fig.\ref{fig:heroArm} B). Here a 2.3 kg payload was bolted to the distal tip of the arm, and the arm was commanded to extend upwards. Extension and a twisting motion were observed. The arm successfully lifted the payload and was able to move through both repeatability test motions (triangle and square) that are discussed in an upcoming section. The arm was also supporting the entire weight of the soft components up against gravity, demonstrating large payload capacity, even when constrained to support its own body weight.

\begin{figure}[t]
    \centering
    \includegraphics[width=\columnwidth]{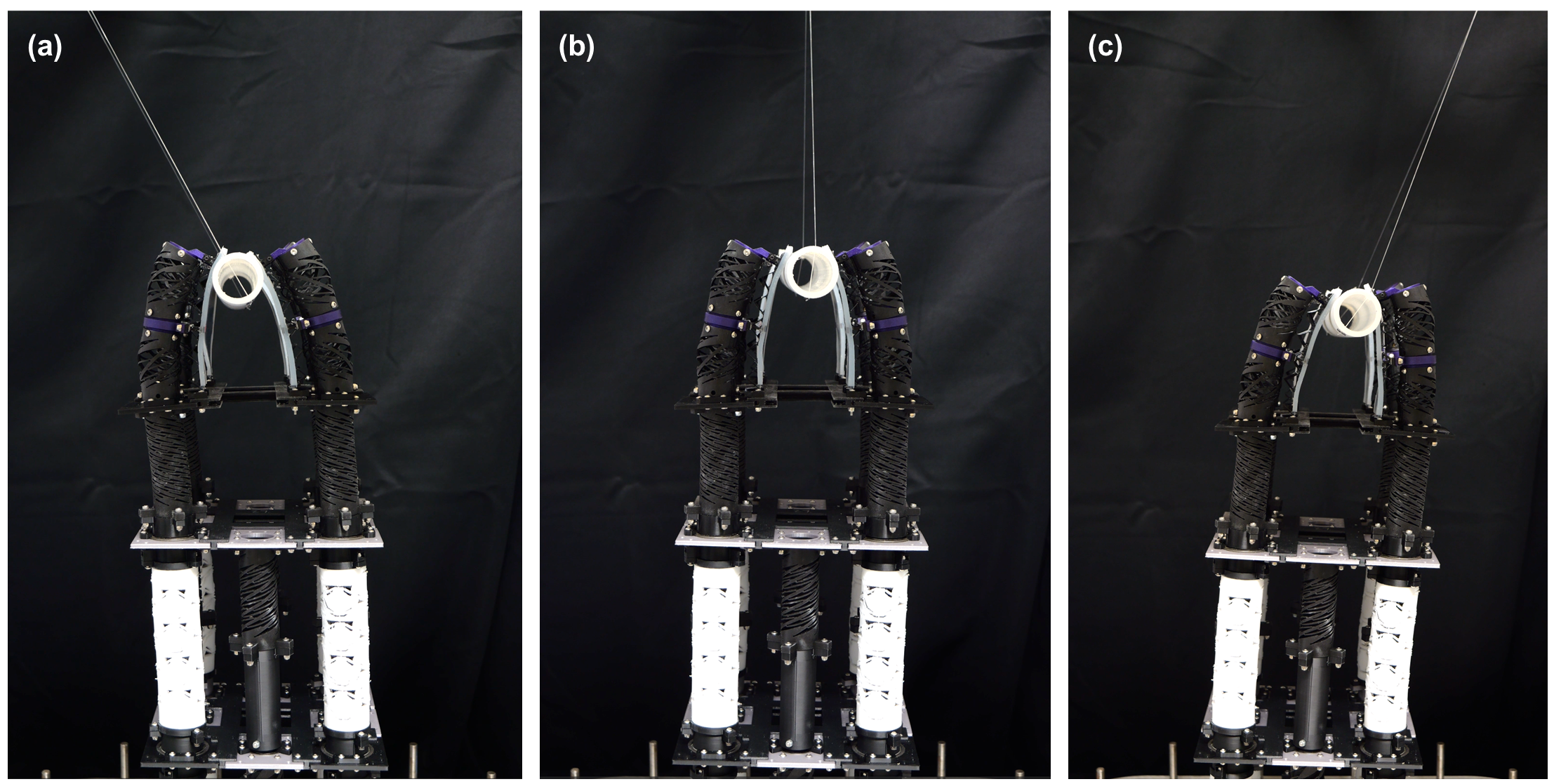}
    \caption[Pull Force Test]{This figure shows the pull force test setup for evaluating the soft HSA arm's active grasping. The robot arm was programmed to fully close the HSA fingers at its most stable position, and a 38.4 mm diameter Teflon tube was pulled out its grasp to measure the maximum force required to dislodge the object. The subfigures (a) and (c) show the test done at the lateral direction, and (b) vertical direction. This test aims to assess the soft robot arm's stability and grasp strength while performing tasks like underwater docking.}
    \label{fig:pullForce}
    %\vspace{-0.3cm}
\end{figure}

 We demonstrated the soft HSA arm's capability to grasp an object by measuring the amount of pull force required to dislodge it from its grasp. First, we replaced the four 0\% HSAs on the fourth layer of the arm with four HSA fingers, and utilized the BETRs to perform active grasping. The test setup can be seen in Fig. \ref{fig:pullForce}. For the test object, we used a 310-mm long Teflon tube with an outer diameter of 38.4 mm. The farther ends of the Teflon tube were connected using strings, and a force gauge was employed to measure the maximum amount of pull force required to remove the object from its grasp. The test was repeated for two different pull directions, vertical, and lateral (left and right). The measured force was then averaged for each direction. The average amount of force required to pull the object out of the arm's grasp in the vertical direction was 20.8 N, and 21.3 N in the lateral direction.

\subsection{Inspection Evaluation} 
%%Workspace%%
\begin{figure*}
    \centering
    \includegraphics[width=0.99\textwidth]{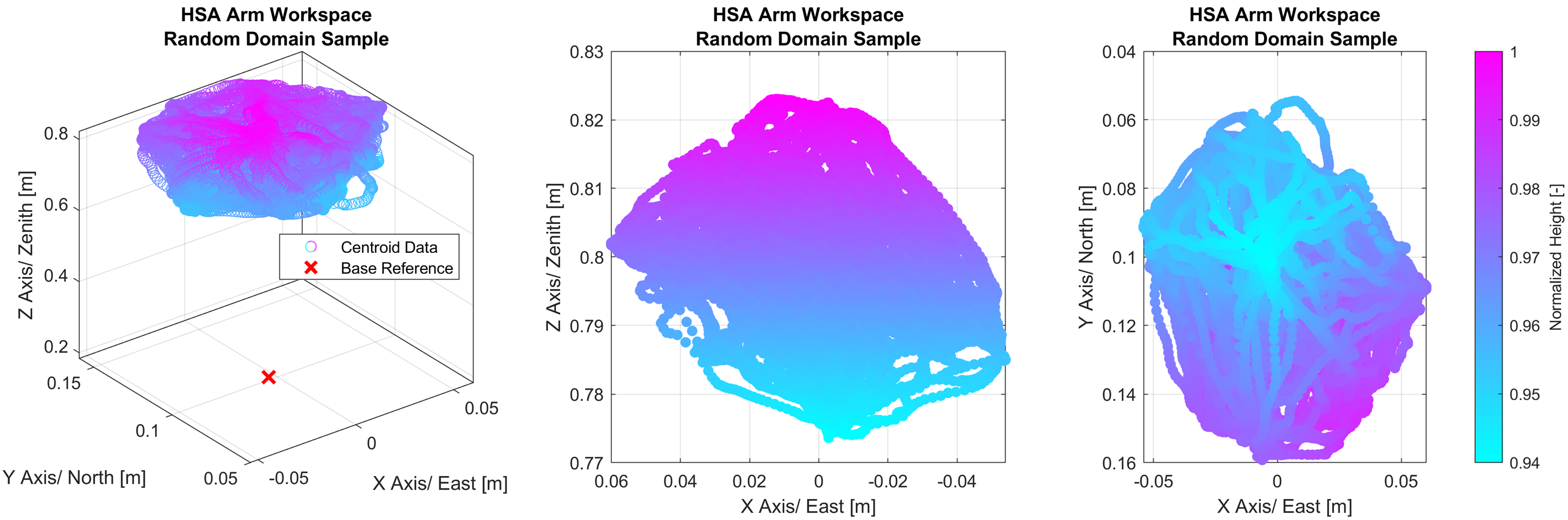}
    \caption[Workspace Analysis]{Here we present the workspace of the Arm as tracked by a rigid body at the tip of distal tip of the robot arm. The arm is moved to the cardinal, inter-cardinal, extended, and twisted positions in a randomized order. The centroid as tracked from the motion capture system is presented in 3D as well as in two slices from the Y and Z direction respectively. The arm covers a range of 106.5 mm in X, 105 mm in Y and 50 mm of extension in Z.
    }
    \label{fig:ArmWorkspace}
    %\vspace{-0.5cm}
\end{figure*}
We performed a mock inspection of a 2 foot (610 mm) diameter pipe, commonly found in the Oil and Gas industry. The goal of the inspection task was to show that the HSA arm can be controlled to move along a curvature, which helps in demonstrating the soft HSA arm's compliance. The HSA arm was placed horizontally before performing the inspection task. A roller was placed on top of each section using FDM 3D-printed mounts to facilitate smooth drag across the surface of the pipe when the arm is in contact. We inspected three common pipe defects, a gash, holes, and scratches \cite{ossai2015pipeline,dai2017PipeAnalysis}. We fixed a Webcam on layer 3 to view and record the pipe defects and arm movement. Highlights from this can be found in Fig. \ref{fig:heroArm} (d) where the holes, gash, and scratches are highlighted in a gold box. The arm was teleoperated in a search pattern from left to right, and top to bottom over the pipe surface. The arm movement was manually stopped when the defect was found, and the search motion was resumed. The video of the HSA arm's inspection task can be found in the supplementary material.

\subsection{HSA Arm Workspace Analysis} \label{armWorkspace}

First, we evaluated the reachable workspace of the HSA arm using random domain sampling among thirteen target positions (rest, two extension, two twist, four cardinals, and four ordinals). The workspace analysis helps in assessing factors such as the arm's ability to reach all the necessary points to perform a specific task. Four tracking markers were attached to 3D-printed mounts, which were fixed to the fourth layer of the HSA arm. A rigid body was then defined using these four markers at the distal tip of the arm and compared to a base reference for tracking. The test was allowed to run continuously for ten minutes to fill in the workspace. The data was recorded using an OptiTrack motion capture system using four Flex 13 cameras. The centroid data of the rigid body at the distal tip was considered to visualize the reachable workspace of the HSA arm, which is shown in Fig. \ref{fig:ArmWorkspace}. Here, the base reference point was determined by subtracting the total height of the arm from the initial position of the tracked centroid. The arm was able to fill in the domain sample quite well with less coverage coming to or from the rest position. The centroid achieved a mobility range of 106.5 mm in X, 105 mm in Y and 50 mm of extension in Z. This provides a reasonable workspace to manipulate heavy objects such as lifting a YCB mustard bottle\cite{YCBObjects} as shown in Fig. \ref{fig:horizontalPayload} (c) and (d). 

\begin{figure*}
    \centering
    \includegraphics[width=0.99\textwidth]{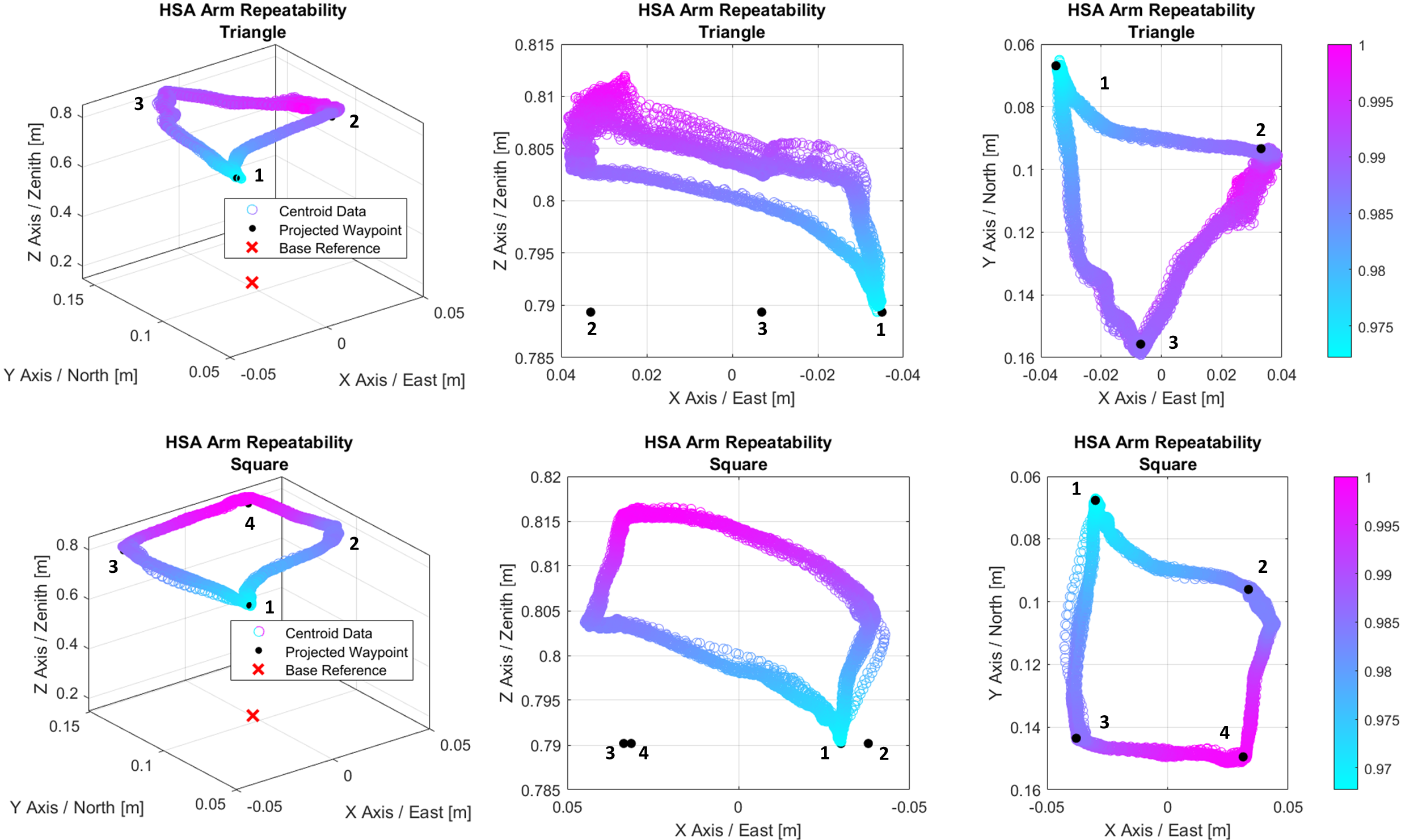}
    \caption[Arm Repeatability]{In this figure we show motor inputs corresponding to a triangular movement pattern (N, SE, SW) and a square movement pattern (NW, NE, SE, SW) repeated 45 times. Color represents Zenith and black dots represent the most common location the arm moved to projected down to the lowest point in its trajectory. As can be observed, the arm is very repeatable, with most variance in position below 5 mm.
    }
    \label{fig:arm-repeatability}
    %\vspace{-0.4cm}
\end{figure*}

\subsection{HSA Arm Repeatability}
Repeatability tests were performed by moving to specific cardinal and inter-cardinal positions. We use triangular and square movement patterns for this evaluation task. The same optical motion tracking setup as described in \ref{armWorkspace} was used to record the data during the movement of our arm. The arm moved to the north (N), southeast (SE), and southwest (SW) positions for a triangular pattern, and to the northwest (NW), northeast (NE), southeast (SE), and southwest (SW) positions for the square movement pattern. The triangular and square patterns were repeated for 45 cycles, where each cycle took  on average 11.6 seconds for the square pattern, and 8.6 seconds for the triangular pattern. The transient startup data from the first cycle was removed. The end effector centroid was tracked and visualized in 3D. For the two repeatability test patterns, the path is shown relative to a reference point which denotes the base of the arm. The base reference points shown in Fig. \ref{fig:arm-repeatability} were calculated by subtracting the total height of the arm (590 mm) from the rest position of the centroid. During the repeatability tests, variances can be observed in some regions (for square pattern: paths 1 to 3 \& 2 to 1; for triangular pattern: paths 3 to 2 \& 2 to 1) which might be caused by the internal PID controllers of the motors. Projected waypoints are shown in black in Fig. \ref{fig:arm-repeatability}. Waypoints were determined by identifying the modal rest position with 1mm bins. The points were then projecting onto the plane where the global minimum Z-value lies so they could be seen. Most X-Y variance between paths falls under 5 mm, demonstrating strong repeatability.

\section{Conclusion and Future Work}

%0.25page
In this paper, we designed and fabricated a soft robot arm using handed shearing auxetics (HSA) and bendable extendable torsionally-rigid flex shafts (BETR). We characterized the 3D force and torque landscape for HSAs in terms of displacement and rotation. We also analyzed the BETRs using mechanical testing to measure its force and torque responses. Then, we performed the workspace and repeatability analysis of the HSA arm using optical motion tracking system. We tested the arm's maximum load capacity and demonstrated the arm's ability in performing tasks such as lifting, docking, and pipe inspection.

The arm was able to lift 2300 g when vertical and more than 600 g when horizontal while also supporting its own weight against gravity. The arm demonstrated a range of motion of 100 mm in translation and 50 mm in extension. Through repeatability tests, the arm showed strong repeatability with most path variance below 5 mm. Future works could consider expanding the available workspace for the arm, and developing more robust controls and controllers.

This work set out to show that soft robot arms can support large payloads and moments, all while supporting their own weight under gravity. This enables more functionality for soft robotic arms by allowing them to be mounted in various configurations. By leveraging the benefits of handed shearing auxetics (HSAs) and bendable, extendable torsionally rigid flex shafts (BETR), torque can be transmitted through a multi-stage arm, capable of lifting and moving with more than 2 kg.
\section*{Acknowledgment}

The authors would like to thank David Leopold, Josh Crumrine, and Joe Torky for their help with 3D printing and assembly of the arm.

\bibliographystyle{IEEEtran}
\bibliography{hsa}

\end{document}